\documentclass{article}

    \usepackage[preprint]{neurips_2018}

\usepackage[utf8]{inputenc} 
\usepackage[T1]{fontenc}    
\usepackage{hyperref}       
\usepackage{url}            
\usepackage{booktabs}       
\usepackage{amsfonts}       
\usepackage{nicefrac}       
\usepackage{microtype}      

\usepackage{lipsum}
\usepackage{amssymb}
\usepackage{amsmath}
\usepackage{graphicx}
\usepackage{listings}
\usepackage{fullpage}
\usepackage{color}
\usepackage[table]{xcolor}
\usepackage{algpseudocode}
\usepackage{algorithm}
\usepackage[normalem]{ulem}
\DeclareMathOperator*{\argmax}{arg\,max}
\DeclareMathOperator*{\argmin}{arg\,min}
\usepackage{tikz}
\usetikzlibrary{fit,positioning}
\usepackage{tikz}
\usetikzlibrary{bayesnet}
\usetikzlibrary{arrows}
\usepackage{color}
\usepackage{graphicx}
\usepackage{caption}
\usepackage{subcaption}
\usetikzlibrary{backgrounds}

\title{Biadversarial Variational Autoencoder}

\author{%
  Arnaud Fickinger \\
  Department of Computer Science\\
  Ecole Polytechnique\\
  Palaiseau, FRANCE \\
  \texttt{arnaud.fickinger@polytechnique.edu} \\
}

\begin{document}

\maketitle

\begin{abstract}
  In the original version of the variational autoencoder \cite{2013arXiv1312.6114K}, Kingma et al. assume Gaussian distributions for the approximate posterior during the inference and for the output during the generative process. This assumptions are good for computational reasons, e.g. we can easily optimize the parameters of a neural network using the reparametrization trick and the KL divergence between two Gaussians can be computed in closed form. However it results in blurry images due to its difficulty to represent multimodal distributions. We show that using two adversarial networks, we can optimize the parameters without any Gaussian assumptions.
\end{abstract}

\section{Introduction}
We want to maximize the evidence lower bound (ELBO) of the marginal likelihood $p_{\theta}(x)$. We can derive the ELBO with the Jensen inequality by marginalizing out the latent variable $z$ and introducing the approximate posterior $q_{\phi}(z|x)$:
\begin{equation}\label{1}
\begin{split}
\log p_{\theta}(x) = & \log\int p_{\theta}(x, z)dz \\
= & \log\int \frac{p_{\theta}(x, z)q_{\phi}(z|x)}{q_{\phi}(z|x)}dz\\
\geq & \mathbb{E}_{q_{\phi}(z|x)}( \log p_{\theta}(x|z) + \log p(z) - \log q_{\phi}(z|x)) \equiv \text{ELBO}
\end{split}
\end{equation}

\section{Inference}
Many works on variational autoencoder assume a Gaussian distribution for the approximate posterior distribution:
\begin{equation}\label{2}
\begin{split}
q_{\phi}(z|x) = \mathcal{N}(z, \mu_{z}, \sigma_{z}^2\mathbf{I})
\end{split}
\end{equation}
where $\mu_{z}$ and $\sigma_{z}$ are neural network functions.

\begin{figure}[H]
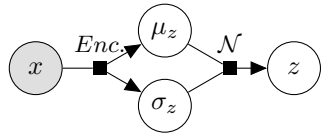

    \centering
    \tikz{
    \node[obs] (y) {$x$};%
    
    \node[latent, right= of y, yshift=0.5cm] (m) {$\mu_{z}$} ; %
      \node[latent, right= of y, yshift=-0.5cm]  (s) {$\sigma_{z}$} ; %
     \node[latent, right= of m, yshift=-0.5cm ] (z) {$z$}; %
     
    
    \factor[right=of y] {enc} {above:$Enc.$} {y} {m,s} ;
     \factor[left=of z] {n} {above:$\mathcal{N}$} {m,s} {z} ;
       }
     \caption{Gaussian Encoder}
\end{figure}

This is convenient for computation but very restrictive for $z$. We introduce an adversarial network that will optimize the parameters of the encoder without the need of any restrictive assumption. To do that, rearrange eq. \eqref{1}:
\begin{equation}\label{3}
\begin{split}
\text{ELBO} \equiv \mathbb{E}_{z\sim q_{\phi}(z|x)}(\log p_{\theta}(x|z)) + \mathbb{E}_{z\sim q_{\phi}(z|x)}(\log p(z) - \log q_{\phi}(z|x))
\end{split}
\end{equation}

The objective being:
\begin{equation}
\begin{split}
\max_{\theta}\max_{\phi}  \, \mathbb{E}_{x \sim \hat{p}_{data}} \text{ELBO}
\end{split}
\end{equation}

where $\phi$ denotes the parameter of the encoder and $\theta$ denotes the parameters of the decoder.

Rewrite the second term of the ELBO in eq. \eqref{3} to bring out a Kullback-Leibler (KL) divergence:
\begin{equation}\label{5}
\begin{split}
& \min_{\theta}\min_{\phi}  \, \mathbb{E}_{x \sim \hat{p}_{data}} \mathbb{E}_{z\sim q_{\phi}(z|x)}(\log q_{\phi}(z|x) - \log p(z))\\
= & \min_{\phi}  \, \mathbb{E}_{x \sim \hat{p}_{data}} KL(q_{\phi}(z|x)||p(z))
\end{split}
\end{equation}

This term corresponds to the KL divergence between the approximate posterior $q_{\phi}(z|x)$ and the prior $p(z)$. Note that it is the reverse KL divergence, ie. the difference between both distributions is bounded by the approximate posterior, which is a better option to learn real modes in case of a multimodal distribution. Inspired by \cite{2016arXiv160600709N}, we define a network with an objective that differs slightly from the original adversarial network \cite{2014arXiv1406.2661G} so the associated generator learns to minimize the reverse KL divergence instead of the Jensen-Shannon divergence. In so doing we are able to optimize the parameters without doing any parametric assumption on the posterior.
Introduce the network $\mathcal{D}_{\phi}:X\times Z \longrightarrow \mathbb{R}$ with the following objective:
\begin{equation}\label{6}
\begin{split}
 \max_{\mathcal{D}_{\phi}} \, V(\mathcal{D}_{\phi},\phi) \equiv \mathbb{E}_{x \sim \hat{p}_{data}}(\mathbb{E}_{z\sim q_{\phi}(z|x)} (1 - \mathcal{D}_{\phi}(x,z)) - \mathbb{E}_{z\sim p(z)} 
( \exp (-\mathcal{D}_{\phi}(x,z))))
\end{split}
\end{equation}
where the parameters $\phi$ is fixed. 

Inspired by \cite{2014arXiv1406.2661G}, write the second term as an integral to find the optimal value of $\mathcal{D}_{\phi}$:

\begin{equation}
\begin{split}
& \mathbb{E}_{x \sim \hat{p}_{data}}(\mathbb{E}_{z\sim q_{\phi}(z|x)} (1 - \mathcal{D}_{\phi}(x,z)) - \mathbb{E}_{z\sim p(z)} 
( \exp (-\mathcal{D}_{\phi}(x,z)))) = \\
&\int \hat{p}_{data}(x)(q_{\phi}(z|x)(1 - \mathcal{D}_{\phi}(x,z)) + p(z) \exp (-\mathcal{D}_{\phi}(x,z))) dzdx
\end{split}
\end{equation}

Given a pair $(a,b)$ in $\mathbb{R}^2$, the function $d\in\mathbb{R} \mapsto a(1-d) - b \exp(-d)$ reaches its maximum at $d^* = \log(\frac{b}{a})$. Hence the maximum of the integral is reached if:

\begin{equation}\label{7}
\begin{split}
&\forall \, x,z, \mathcal{D}_{\phi}^*(x, z) = \log(\frac{p(z)}{q_{\phi}(z|x)}) \\
\end{split}
\end{equation}

By replacing eq. \eqref{7} in eq. \eqref{6}, we show that the optimal value function $V(\mathcal{D}^*_{\theta,\theta},\theta,\theta)$ reached by the discriminator, the generator being fixed, is the KL divergence in eq. \eqref{5}:
\begin{equation}
\begin{split}
V(\mathcal{D}^*_{\theta,\theta},\theta,\theta) = & \mathbb{E}_{x \sim \hat{p}_{data}}(\mathbb{E}_{z\sim q_{\phi}(z|x)} (1 - \log(\frac{p(z)}{q_{\phi}(z|x)})) \\
& - \mathbb{E}_{z\sim p(z)} 
( \exp (-\log(\frac{p(z)}{q_{\phi}(z|x)})))) \\
= & \mathbb{E}_{x \sim \hat{p}_{data}}(\mathbb{E}_{z\sim q_{\phi}(z|x)} (\log(\frac{q_{\phi}(z|x)}{p(z)})) \\
= & \mathbb{E}_{x \sim \hat{p}_{data}} KL(q_{\phi}(z|x)||p(z))
\end{split}
\end{equation}

In so doing we can optimize the second term of the ELBO with a minimax game with value function $V(\mathcal{D}_{\phi},\phi)$:
\begin{equation}
\begin{split}
\min_{\phi}  \, \max_{\mathcal{D}_{\phi}} \, V(\mathcal{D}_{\phi},\phi)
\end{split}
\end{equation}

\begin{figure}[H]
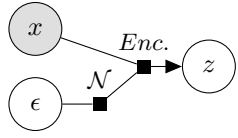

    \centering
    \tikz{
    \node[obs, yshift=0.5cm] (y) {$x$};%
    \node[latent, yshift=-0.5cm] (e) {$\epsilon$};%
     \node[latent, right= of enc, yshift=0cm ] (z) {$z$}; %
     
    
    \factor[right=of e] {n0} {above:$\mathcal{N}$} {e} {} ;
    \factor[right=of n0, yshift=0.5cm] {enc} {above:$Enc.$} {n0,y} {z} ;
       }
     \caption{Our Encoder}
\end{figure}

\section{Generative process}

Many works on variational autoencoder assume also a Gaussian distribution for the output distribution:

\begin{equation}
\begin{split}
p(x|z) = & \mathcal{N}(x|\mu, \sigma^2\mathbf{I}) \\
= & \frac{1}{(2\pi)^{n/2}\sigma}\exp (-\frac{||x-\mu||_2^2}{2\sigma^2})
\end{split}
\end{equation}

where $\mu$ is a neural network function and $\sigma^2$ is a hyperparameter.

The negative log likelihood of this distribution is an affine function of the L2 norm, hence we often encounter a L2 reconstruction term in works on variational autoencoders :
\begin{equation}
\begin{split}
-\log p(x|z) = & \log ((2\pi)^{n/2}\sigma) +  \frac{||x-\mu||_2^2}{2\sigma^2} 
\end{split}
\end{equation}

\begin{figure}[H]
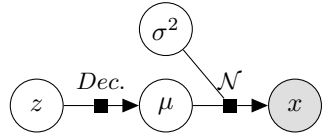

    \centering
    \tikz{
    
     \node[latent] (z) {$z$}; %
     \node[latent, right= of z ] (g) {$\mu$};
     \node[latent, above= of g, yshift=-0.7cm] (s) {$\sigma^2$};
     \node[obs,right= of g ] (x) {$x$};
    
     \factor[right=of z] {dec} {above:$Dec.$} {z} {g} ;
     \factor[right=of g] {n2} {above:$\mathcal{N}$} {g, s} {x} ;
     
      }
     \caption{Gaussian Decoder}
\end{figure}

Rearrange the first term of the objective in eq. \eqref{3} to bring out a direct KL divergence:
\begin{equation}\label{12}
\begin{split}
& \argmax_{\theta}\argmax_{\phi}  \, \mathbb{E}_{x \sim \hat{p}_{data}} \mathbb{E}_{z\sim q_{\phi}(z|x)}(\log p_{\theta}(x|z)) \\
= & \argmin_{\theta}\argmin_{\phi} \mathbb{E}_{z\sim q_{\phi}(z|x)} \mathbb{E}_{x \sim \hat{p}_{data}} (\log \hat{p}_{data}(x) - \log p_{\theta}(x|z)) \\
= & \argmin_{\theta}\argmin_{\phi} \mathbb{E}_{z\sim q_{\phi}(z|x)} KL(\hat{p}_{data}(x)||p_{\theta}(x|z))
\end{split}
\end{equation}

This time we choose an adversarial objective so that the associated generator learns to minimize the direct KL divergence. Introduce the network $\mathcal{D}_{\theta,\phi}:X\times Z \longrightarrow \mathbb{R}$ with the following objective:
\begin{equation}\label{13}
\begin{split}
 \max_{\mathcal{D}_{\theta,\phi}} \, V(\mathcal{D}_{\theta,\phi},\theta,\phi) \equiv \mathbb{E}_{z\sim q_{\phi}(z|x)}(\mathbb{E}_{x \sim \hat{p}_{data}} (\mathcal{D}_{\theta,\phi}(x,z)) - \mathbb{E}_{x\sim p_{\theta}(x|z)} 
( \exp (\mathcal{D}_{\theta,\phi}(x,z)-1)))
\end{split}
\end{equation}
where the parameters $\theta$ and $\phi$ are fixed. 

Write the second term as an integral to find the optimal value of $\mathcal{D}_{\theta,\phi}$:

\begin{equation}
\begin{split}
& \mathbb{E}_{z\sim q_{\phi}(z|x)}(\mathbb{E}_{x \sim \hat{p}_{data}} (\mathcal{D}_{\theta,\phi}(x,z)) - \mathbb{E}_{x\sim p_{\theta}(x|z)} 
( \exp (\mathcal{D}_{\theta,\phi}(x,z)-1))) = \\
&\int q_{\phi}(z|x) (\hat{p}_{data}(x)( \mathcal{D}_{\theta,\phi}(x,z)) + p_{\theta}(x|z) \exp (\mathcal{D}_{\theta,\phi}(x,z)-1)) dxdz
\end{split}
\end{equation}

Given a pair $(a,b)$ in $\mathbb{R}^2$, the function $d\in\mathbb{R} \mapsto ad - b \exp(d-1)$ reaches its maximum at $d^* = 1+ \log(\frac{a}{b})$. Hence the maximum of the integral is reached if:

\begin{equation}\label{15}
\begin{split}
&\forall \, x,z, \mathcal{D}_{\theta,\phi}^*(x, z) = 1+ \log(\frac{\hat{p}_{data}(x)}{p_{\theta}(x|z)}) \\
\end{split}
\end{equation}

By replacing eq. \eqref{15} in eq. \eqref{13}, we show that the optimal value function $V(\mathcal{D}^*_{\theta,\phi},\theta,\phi)$ reached by the discriminator, the generator being fixed, is the direct KL divergence in eq. \eqref{12}:
\begin{equation}
\begin{split}
V(\mathcal{D}^*_{\theta,\phi},\theta,\phi) = & \mathbb{E}_{z\sim q_{\phi}(z|x)}(\mathbb{E}_{x \sim \hat{p}_{data}} (1+ \log(\frac{\hat{p}_{data}(x)}{p_{\theta}(x|z)})) - \\
& \mathbb{E}_{x\sim p_{\theta}(x|z)} 
( \exp (1+ \log(\frac{\hat{p}_{data}(x)}{p_{\theta}(x|z)})-1))) \\
= & \mathbb{E}_{z\sim q_{\phi}(z|x)}(\mathbb{E}_{x \sim \hat{p}_{data}} ( \log(\frac{\hat{p}_{data}(x)}{p_{\theta}(x|z)})) \\
= & \mathbb{E}_{z\sim q_{\phi}} KL(\hat{p}_{data}(x)||p_{\theta}(x|z))
\end{split}
\end{equation}

In so doing we can optimize the second term of the ELBO with a minimax game with value function $V(\mathcal{D}_{\theta,\phi},\theta,\phi)$:
\begin{equation}
\begin{split}
\min_{\theta}\min_{\phi}  \, \max_{\mathcal{D}_{\theta,\phi}} \, V(\mathcal{D}_{\theta,\phi},\theta,\phi)
\end{split}
\end{equation}

\begin{figure}[H]
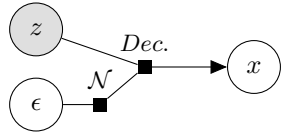

    \centering
    \tikz{
    \node[obs, yshift=0.5cm] (y) {$z$};%
    \node[latent, yshift=-0.5cm] (e) {$\epsilon$};%
     \node[latent, right= of enc, yshift=0cm ] (z) {$x$}; %
     
    
    \factor[right=of e] {n0} {above:$\mathcal{N}$} {e} {} ;
    \factor[right=of n0, yshift=0.5cm] {enc} {above:$Dec.$} {n0,y} {z} ;
       }
     \caption{Our Decoder}
\end{figure}

Finally we have transformed the optimization of the ELBO into a minimax game involving two discriminators:
\begin{equation}
\begin{split}
\min_{\theta}\min_{\phi}  \, (\max_{\mathcal{D}_{\theta,\phi}} \, V(\mathcal{D}_{\theta,\phi},\theta,\phi)+\max_{\mathcal{D}_{\phi}} \, V(\mathcal{D}_{\phi},\phi))
\end{split}
\end{equation}

\section{Implementation}
The model is implemented with PyTorch. The implementation is available here: \\ \href{https://github.com/ArnaudFickinger/BAVAE}{\texttt{https://github.com/ArnaudFickinger/BAVAE}}.







\end{document}